\documentclass[10pt, conference, letterpaper]{IEEEtran}
\IEEEoverridecommandlockouts
\usepackage{subfigure}
\usepackage[utf8]{inputenc}
\usepackage[style=ieee]{biblatex}
\usepackage{amsmath,amssymb,amsfonts}
\usepackage{algorithmic}
\usepackage{graphicx}
\usepackage{textcomp}
\usepackage{float}
\usepackage{amsmath}
\usepackage{mathtools}
\usepackage{xcolor}
\usepackage{physics}
\usepackage[linesnumbered,ruled,vlined]{algorithm2e}
\usepackage{booktabs}
\usepackage{multirow}
\usepackage{makecell}
\usepackage{footnote}
\usepackage{longtable}
\usepackage{threeparttable}
\usepackage{graphicx}
\usepackage{color}
\usepackage{caption}

\usepackage{xcolor}

\newcommand{\er}{
    \color{black}
    \captionsetup{font={color=black}}
}

\DeclareUnicodeCharacter{1EF3}{\`y}
\usepackage{etoolbox}
\makeatletter
\patchcmd{\@makecaption}
  {\scshape}
  {}
  {}
  {}
\makeatother

\newcommand\numeq[1]%
  {\stackrel{\scriptscriptstyle(\mkern-1.5mu#1\mkern-1.5mu)}{=}}
\newcommand\numleq[1]%
  {\stackrel{\scriptscriptstyle(\mkern-1.5mu#1\mkern-1.5mu)}{\leq}}
  
\newcommand\numgeq[1]%
  {\stackrel{\scriptscriptstyle(\mkern-1.5mu#1\mkern-1.5mu)}{\geq}}
  
\usepackage{amsthm}
\theoremstyle{definition}
\newtheorem{theorem}{\textbf{Theorem}}
\newtheorem{lemma}{Lemma}
\newtheorem{remark}{Remark}
 
\begin{document}

\title{\(\gamma\)-FedHT: Stepsize-Aware Hard-Threshold Gradient Compression
in Federated Learning \\
}

\author{
\IEEEauthorblockN{
Rongwei Lu\IEEEauthorrefmark{2}, 
Yutong Jiang\IEEEauthorrefmark{2},
Jinrui Zhang\IEEEauthorrefmark{2},
Chunyang Li\IEEEauthorrefmark{3}\IEEEauthorrefmark{1}\thanks{\IEEEauthorrefmark{1} Chunyang Li has been pre-admitted to Tsinghua Shenzhen International Graduate School when doing this work.},
Yifei Zhu\IEEEauthorrefmark{4}
Bin Chen\IEEEauthorrefmark{3},
Zhi Wang\IEEEauthorrefmark{2}\IEEEauthorrefmark{7}\thanks{\IEEEauthorrefmark{7}Corresponding author.}} 
\IEEEauthorblockA{\IEEEauthorrefmark{2}Tsinghua Shenzhen International Graduate School, Tsinghua University}
\IEEEauthorblockA{\IEEEauthorrefmark{3}Harbin Institute of Technology, Shenzhen}
\IEEEauthorblockA{\IEEEauthorrefmark{4}UM-SJTU Joint Institute, Shanghai Jiao
Tong University}

\IEEEauthorblockA{
\{lurw24, jiang-yt24,  zhangjr23\} @mails.tsinghua.edu.cn,
210110812@stu.hit.edu.cn,\\
yifei.zhu@sjtu.edu.cn,
chenbin2021@hit.edu.cn, 
wangzhi@sz.tsinghua.edu.cn}
}

\maketitle

\begin{abstract}

Gradient compression can effectively alleviate communication bottlenecks in Federated Learning (FL). Contemporary state-of-the-art sparse compressors, such as Top-$k$, exhibit high computational complexity, up to $\mathcal{O}(d\log_2{k})$, where $d$ is the number of model parameters. The hard-threshold compressor, which simply transmits elements with absolute values higher than a fixed threshold, is thus proposed to reduce the complexity to $\mathcal{O}(d)$. 
However, the hard-threshold compression causes accuracy degradation in FL, where the datasets are non-IID and the stepsize $\gamma$ is decreasing for model convergence. The decaying stepsize reduces the updates and causes the compression ratio of the hard-threshold compression to drop rapidly to an aggressive ratio. At or below this ratio, the model accuracy has been observed to degrade severely. To address this, we propose $\gamma$-FedHT, a stepsize-aware low-cost compressor with Error-Feedback to guarantee convergence. Given that the traditional theoretical framework of FL does not consider Error-Feedback, we introduce the fundamental conversation of Error-Feedback. We prove that $\gamma$-FedHT has the convergence rate of $\mathcal{O}(\frac{1}{T})$ ($T$ representing total training iterations) under $\mu$-strongly convex cases and $\mathcal{O}(\frac{1}{\sqrt{T}})$ under non-convex cases, \textit{same as FedAVG}. Extensive experiments demonstrate that $\gamma$-FedHT improves accuracy by up to $7.42$\% over Top-$k$ under equal communication traffic on various non-IID image datasets.

\end{abstract}


\section{Introduction} \label{introduction}

FL is an increasingly important Distributed Machine Learning (DML) framework that addresses the critical need for data privacy in model training across multiple edge nodes \cite{FedSurvey, FedSurvy2}.  FL requires a decaying stepsize\footnote{
We focus on the original FedAVG \cite{fedavg} with a single learning rate on the client side instead of two on both the server and client sides.
}, not a fixed one, to ensure model convergence in non-IID scenarios \cite{li2019convergence}, which are common in FL \cite{niid2020icml}. In FL, gradient compression has been widely adopted to alleviate communication bottlenecks. The classic process of FL training with gradient compression involves three main steps: (1) clients train the local model for several iterations to obtain updates; (2) clients compress the updates and send them to the central server; (3) the server decompresses the updates, aggregates them (\textit{e.g.}, averaging them in FedAVG \cite{fedavg}), and updates the global model.
Gradient compression methods can be classified into three categories: (1) sparsification, transmitting a part of the gradients; (2) quantization, mapping high-precision elements into low-precision ones; and (3) low-rank, decomposing the gradient into two low-rank matrices. Sparsification compression is often preferred due to its superior efficiency in reducing redundant gradient information. The gradient sparsification usually comes with Error-Feedback (EF) \cite{docofl2023icml,stich2019error}, a popular mechanism that collects and reuses the errors from the gradient compression to mitigate the compression bias and guarantee convergence. \er The two popular sparsification compressors are the Top-$k$ compressor \cite{stich2020communication} and the hard-threshold compressor \cite{hardthreshold}. The Top-$k$ compressor transmits elements with the top $k$ absolute values, while the hard-threshold compressor transmits elements with absolute values larger than a threshold.

Top-$k$ is recognized as the state-of-the-art (SOTA) sparsification compressor in FL\footnote{Many well-performing hybrid gradient compressors in FL use Top-$k$ for sparsification \cite{cui2021slashing, FedZIP,stc}.}, but its counterpart, the hard-threshold compressor, is not suitable for FL. \textit{We are the first to demonstrate that the hard-threshold compressor shows inferior convergence rates compared to Top-$k$ in FL} (as shown in Fig.~\ref{motivate:figa} in Sec.~\ref{measurement 1}) by the control variable method. This is because the hard-threshold compression is sensitive to the combination of the decaying stepsize and non-IID cases. In particular,
\textit{we examine the sensitivity of the hard-threshold compression to such cases} (as shown in Fig.~\ref{motivate fig b} in Sec.~\ref{measurement 2}) through the full factorial experimental design \cite{fullfactorial}, an approach involving systematically varying all experimental factors and their combinations to understand their effects comprehensively. The compression ratio of the hard-threshold compressor drops rapidly to an extremely aggressive value (like $0.1\%$ for a CNN model with CIFAR-10 datasets \cite{aji2017sparse}) in the combination of the decaying stepsize and non-IID cases. Such aggressive compression stops the model from converging in non-IID scenarios, thereby reducing the model accuracy. 

Although the hard-threshold compression degrades the accuracy, the lightweight compression is extremely appealing in FL. In fact, the computation cost of Top-$k$ is up to hundreds of times of the hard-threshold compression\footnote{According to Fig.~15(d) in the appendix of the previous work \cite{SIDCo}, the compression time of Top-$k$ is hundreds of times that of SIDCo. Furthermore, the absolute compressor is more effective than SIDCo.}, primarily due to two reasons: (1) Top-$k$ selection has a time complexity of $O(d\log_2{k})$, and $\log_2{k}$ depends on the model scale (\textit{e.g.}, it can be up to nearly $30$ for GPT2 \cite{gpt2}), while the hard-threshold compression requires traversal of parameters with the time complexity of $O(d)$; and (2) Top-$k$ selection does not perform well on accelerators such as GPUs \cite{SIDCo}. Clearly, the ideal sparsification compressor in FL is one that has both the low-cost computation of the hard-threshold compression and the superior performance of Top-$k$. \textit{Sadly, no sparsification compressor in FL has been developed to have a time complexity of $O(d)$ and the same theoretical convergence rate as vanilla FedAVG so far.}

We propose $\gamma$-FedHT (as shown in Algo.~\ref{r-FedHT} in Sec.~\ref{section r-fedht}), an ideal sparsification compressor in FL satisfying the above properties. $\gamma$-FedHT is a stepsize-aware hard-threshold compressor with vanilla EF, avoiding the accelerator-unfriendly operations like Top-$k$ selection, and inheriting the low-cost property. To improve the performance, the threshold should satisfy the increasing and then decreasing monotonicity with a limit of zero. Combining two simple functions, the inverse proportional function and the logarithmic function, the adaptive threshold can satisfy the two mathematical properties without introducing more hyperparameters. Although there have been efforts to theoretically validate gradient compression algorithms in FL \cite{cui2021slashing,cui2022infocom}, these works have not considered EF, which is important and necessary for sparsification compression. To derive the convergence rate of our design, we solve the problem of \textit{how to integrate gradient compression with EF into the theoretical framework of FL}. We fuse the mathematical description of EF into the framework and establish an iterative equation. Based on this, we derive the convergence rates. The convergence rates of $\gamma$-FedHT are $\mathcal{O}(\frac{1}{T})$ under $\mu$-strongly convex functions and $\mathcal{O}(\frac{1}{\sqrt{T}})$ under non-convex functions, \textit{the same rate as FedAVG without compression}. 

Our contributions are as follows:

\begin{itemize}
    \item We are the first to reveal that the model trained with the hard-threshold compression converges less effectively than the one trained with Top-$k$ compressor by the controlled variable method. We use a full factorial experimental design to demonstrate that it is the combination of the decaying stepsize and non-IID scenarios that contributes to the failure of the hard-threshold compression in FL.

    \item We propose $\gamma$-FedHT, the first sparsification compressor in FL with a time complexity of $\mathcal{O}(d)$ and the same convergence rate as vanilla FedAVG. We expand the application of the traditional FL theoretical framework and derive the convergence rate of FedAVG with gradient compression and EF, based on introducing the iterative equation of EF.    
    
    \item We apply $\gamma$-FedHT to both real-world non-IID and artificially partitioned non-IID datasets, including convex cases (\textit{e.g.} Logistic) and non-convex cases (\textit{e.g.} VGG, CNN and GPT2).
    The experimental results validate the great compression-accuracy trade-offs of our design. Under equal traffic communication, $\gamma$-FedHT can improve accuracy by up to $7.42$\% over Top-$k$ on the CNN model with non-IID datasets.
\er 
\end{itemize}


\section{Preliminaries} \label{II}

Our research focuses on the synchronous FedAVG algorithm with the sparsification compressor. We aim to provide a succinct overview of the optimization problem of FedAVG, the differences between FL and traditional DML, gradient compression, and Error-Feedback, with a particular emphasis on the hard-threshold compressor.

\noindent \textbf{The optimization problem of FL:}
the optimization problem of FL minimizes a loss function $f$ as follows:
\begin{equation}
\underset{\textbf{x}}{\min}\left[ f(\textbf{x}):=\sum_{i=1}^n p_i f_i(\textbf{x})\right] , \nonumber
\end{equation}
where $n$ represents the number of clients. The $i$-th client possesses a mutually disjoint partition $D^i$ of the overall 
 training dataset $D$ and the training weight $p_i = \dfrac{|D^i|}{\sum_{i'=1}^n |D^{i'}|}$. The local training target $f_i(\textbf{x})$ is the loss function evaluated on $D^i$. 

\noindent \textbf{FL vs. Traditional DML:}
FL distinguishes itself from traditional DML in the following four fundamental aspects:
\begin{itemize}
    \item \textit{Non-IID:} In FL, to safeguard data privacy, datasets cannot be exchanged between nodes, resulting in unbalanced data distributions and quantities across nodes \cite{li2019convergence}. This contrasts with traditional DML, where datasets are uniformly partitioned across nodes, typically yielding IID datasets.

    \item \textit{Decaying-$\gamma$:} FL necessitates the stepsize $\gamma$ that decays to zero to guarantee model convergence \cite{niid2020icml}. Fixed-$\gamma$ in FL can cause significant deviation of the global model from the optimal, with the L2 norm of the difference being proportionate to $\gamma^2$. However, in traditional DML, decaying-$\gamma$ is not necessary. 
    \item \textit{Infrequent communication:} FL is characterized by low bandwidth and high latency due to the training across WAN \cite{DGA}. This necessitates infrequent communication, where aggregation occurs after several training iterations, not after each one. This study adheres to a fixed communication frequency \(E\), consistent with vanilla FedAVG, and does not explore adaptive strategies for \(E\).

    \item \textit{Partial node participation:} Due to device heterogeneity and unreliability in FL, aggregation rounds typically involve only the fastest-responding nodes, avoiding delays from slower ones. In contrast, DML benefits from homogeneous and reliable nodes, allowing consistent participation in all aggregation rounds. This paper focuses on the strategy of uniform random node selection \cite{li2019convergence}.
  
\end{itemize}

\noindent \textbf{Gradient compression:} According to whether the mathematical expectation of the gradient changes before and after compression \cite{stich2020communication}, compressors can be classified into biased and unbiased compressors. The biased compressors with EF \cite{wu2018error} are widely used in FL because they can apply more aggressive compression than unbiased compressors.

\noindent \textbf{The hard-threshold compressor:} The hard-threshold algorithm, also known as the threshold-$\lambda$ sparsification compressor, is a variant of Top-$k$, with which it shares a certain conversion relationship: a given $k$ corresponds to a specific $\lambda$. With the objective of minimizing the sum of compression errors throughout the entire training, the hard-threshold achieves a more favorable compression-accuracy trade-off than Top-$k$ \cite{hardthreshold}.
The hard-threshold compressor is also called the absolute compressor due to its key property that the error term possesses an upper bound independent of $\textbf{x}$. We denote the threshold as $\lambda$ and the absolute compressor as $\textbf{C}_{\lambda}(\cdot)$. $\textbf{C}_{\lambda}(\cdot)$ represents a mapping: $\mathbb{R}^d \rightarrow \mathbb{R}^d$, characterized by the following property:
\begin{equation}
   \mathbb{E}_{\textbf{C}_{\lambda}}\lVert \textbf{C}_{\lambda}(\textbf{x}) - \textbf{x} \rVert^2 \leq d \gamma^2 \lambda^2 . \nonumber
\end{equation}

We focus on FedAVG with the hard-threshold compression, which has not been investigated in FL.

\noindent \textbf{Error-Feedback:} The error-feedback mechanism \cite{stich2019error} in gradient compression involves storing and accumulating compression errors over iterations, which are then added to the gradient in subsequent steps to ensure accurate gradient updates and mitigate the loss of information due to compression \cite{alistarh2018convergence,1-bitSGD,EFQ,Hogwild}. Some works propose new 
EF mechanisms to guarantee sharper convergence rate, like EF21 \cite{EF21} and EControl \cite{Stich2023EControl}, by introducing momentum terms or other compensation terms. We focus on vanilla EF for two reasons: (1) EF is orthogonal to sparsification compressors (\textit{e.g.}, EF and EF21 are orthogonal to sparsification); and (2) New EF mechanisms tend to introduce hyperparameters or require additional storage space, complicating the optimization problem (\textit{e.g.}, the performance of EControl is sensitive to the hyperparameter).

\section{Motivation} \label{III}
\begin{figure}[!t]
\centering
\includegraphics[width=0.98\linewidth]{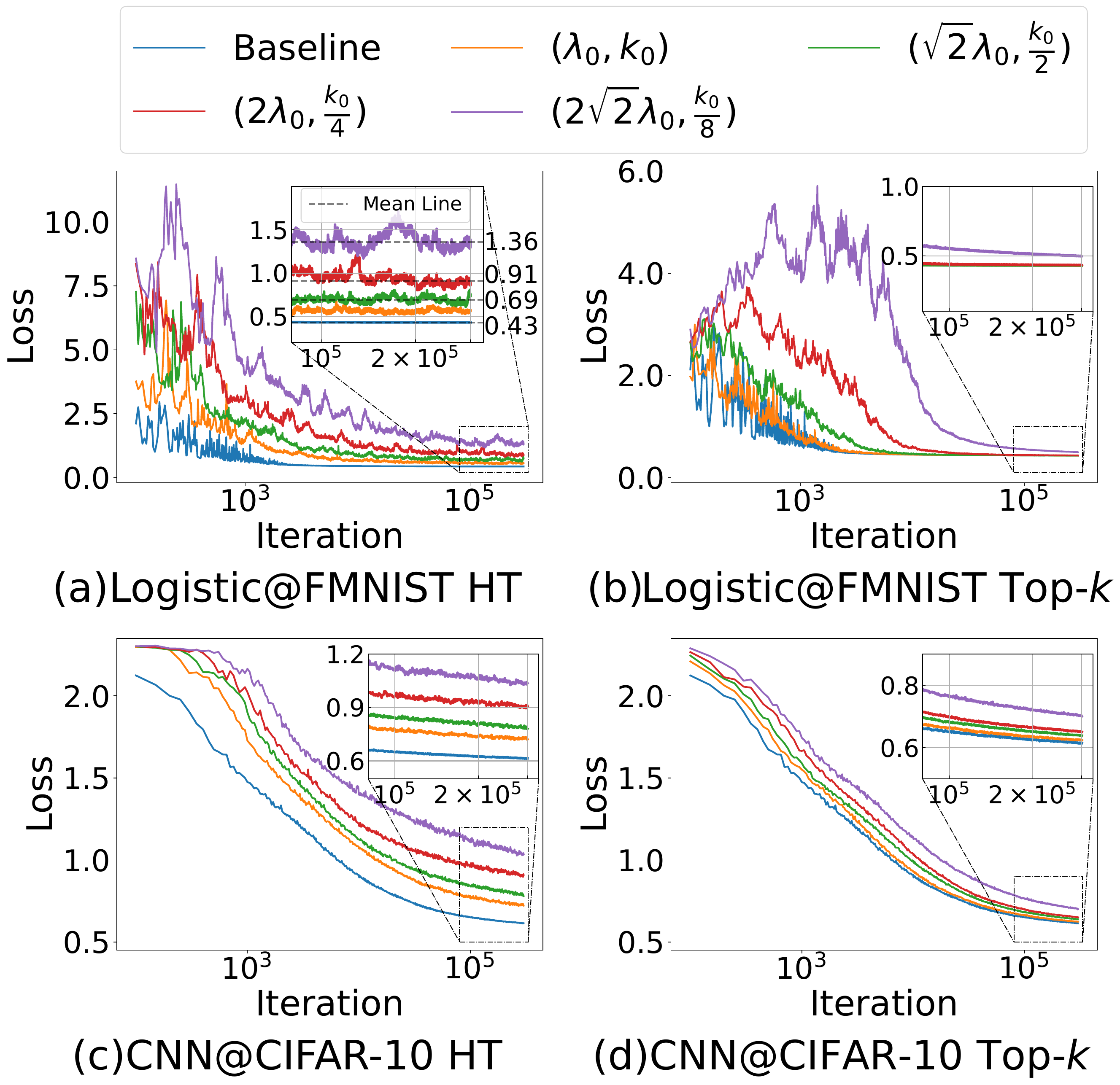}
\caption{The global loss curves (Loss vs. Iterations) for FedAVG with hard-threshold compression (denoted as HT, left) and Top-$k$ (right) on different tasks (top to bottom). $\lambda_0$ is  $\frac{\sqrt{2}}{2}$ (as well as $\frac{\sqrt{2}}{10}$) in Logistic@FMNIST (CNN@CIFAR-10). $k_0$ is $1\%$ for two tasks. The loss curves in (a, c) do not converge to the baseline, while those in (b, d) converge.}
\label{motivate:figa}
\vspace{-1em}
\end{figure}

\begin{figure}[!t]
\centering
\includegraphics[scale=0.088]{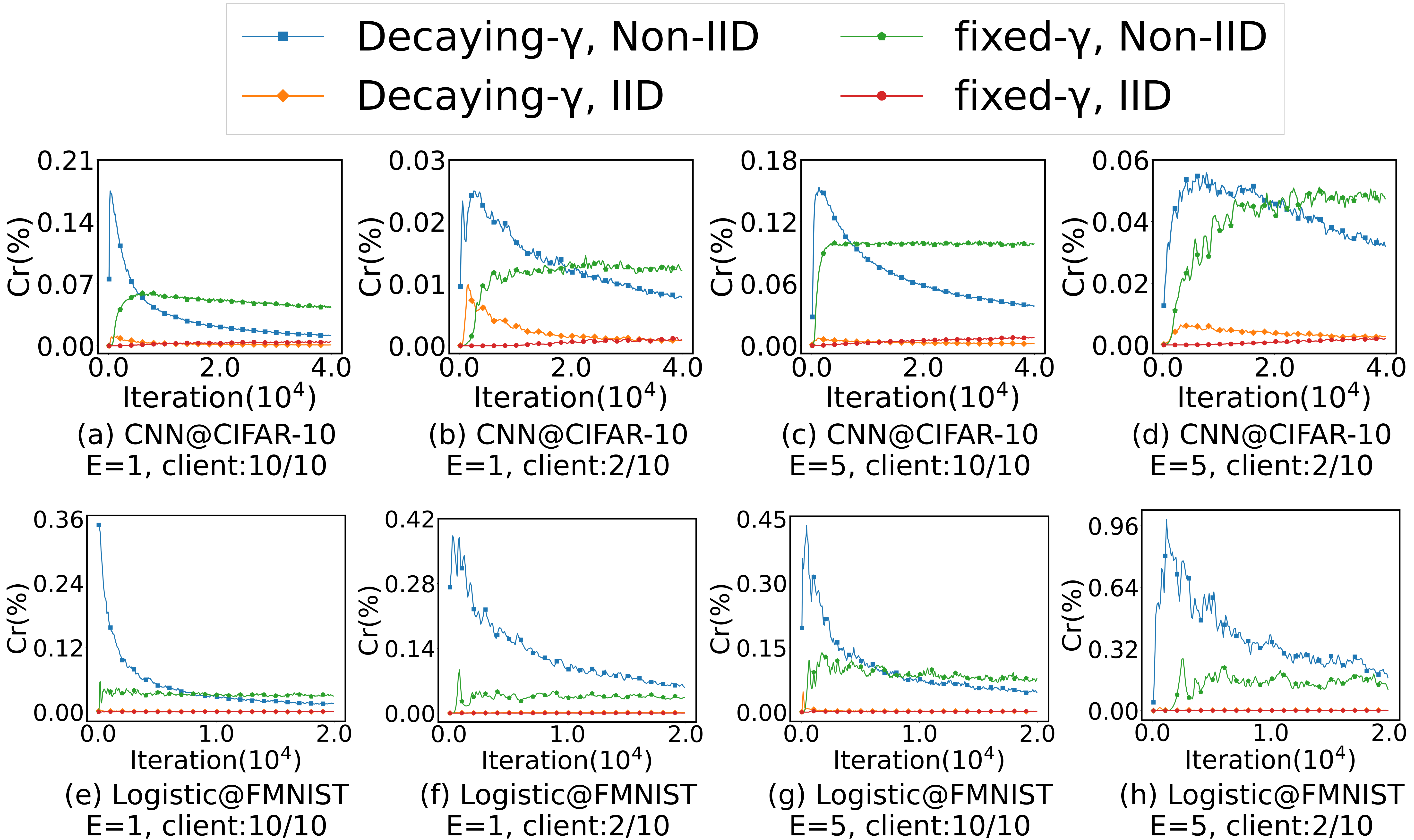}
\caption{Compression curves (Compression ratio vs. Iterations) for HT with $\lambda=1$ under different settings in the convex (a-d) and non-convex (e-h) cases. Here, $E=1$ and  $E=5$ denote frequent and infrequent communication, respectively, while Client: $10/10$ and $2/10$ indicate partial and full partition scenarios. Each subplot consistently demonstrates that in non-IID scenarios, the decaying-$\gamma$ prompts the hard-threshold compressor to  engage in increasingly aggressive compression strategies as training progresses into the late stage.}
\label{motivate fig b}
\vspace{-1em}
\end{figure}

This measurement answers the following questions:
\begin{itemize}
    \item Does the hard-threshold compressor induce non-convergence in FL? If so, is the degree of non-convergence correlated with $\lambda$? (Fig.~\ref{motivate:figa} in Sec.~\ref{measurement 1})

    \item What are the underlying causes for the non-convergence exhibited by the hard-threshold compressor in FL? (Fig.~\ref{motivate fig b} in Sec.~\ref{measurement 2}) 
\end{itemize}

Specifically, we employ two measurements to validate our motivation thoroughly. 1) Logistic Regression \cites{stc} on Fashion-MNIST \cite{fmnist} dataset (denoted as Logistic@FMNIST), which is the most classical convex case \cite{cui2022infocom}. 2) CNN on CIFAR-10 \cite{cifar10} dataset (denoted as CNN@CIFAR-10), a widely-used non-convex scenario \cite{cui2022infocom}.
To align with previous works \cite{dagc,li2019convergence}, we set the number of clients as $10$, the communication frequency $E = 5$ (\textit{i.e.}, global communication after every $5$ local iterations). We denote the stepsize at the $t$-th iteration as $\gamma_t$
and set $\gamma_t = \frac{100}{t+1000}$. The non-IID partition strategy for Logistic@FMNIST (as well as CNN@CIFAR-10) is $\#C=2$ ($\#C=5$) quantity-based label imbalance \cite{niidbench}, where $\#C=2$ (as well as $\#C=5$) means that each node owns data samples of $2$ ($5$) labels and there
is no overlap between the samples of each partition. We set vanilla FedAVG as the baseline.

\subsection{Poor Convergence of the Hard-Threshold Compression} \label{measurement 1}

In Fig.~\ref{motivate:figa}, the comparative analysis
of Top-$k$  and the hard-threshold compression reveals that the hard-threshold compression makes the global model far from the optimal model. Specifically, the loss curves of Top-$k$ (b, d) converge to the baseline, while ones of hard-threshold compression (a, c) does not converge to the baseline. This discrepancy indicates that the \textit{hard-threshold compressor introduces the accuracy degradation and the convergence rate of FedAVG with hard-threshold compression cannot reach that of vanilla FedAVG}, both for the convex case and the non-convex one.

\textit{The l2-norm between the global model parameters and the optimal model is positively correlated with $\lambda$.} In Fig.~\ref{motivate:figa}(a) and \ref{motivate:figa}(c), we observe that the larger the value of $\lambda$, the greater the divergence between the loss of FedAVG with the hard-threshold compressor and the baseline loss. This indicates an increasingly wider gap between the performance of the global model and optimal model\footnote{This is referring to Appendix ~C-3 in \cite{li2019convergence}. We consider the global model to have converged to the optimal model when its loss matches the baseline loss. Otherwise, there is a distance between the global and optimal models.}. Moreover, in convex scenarios, we observed a positive correlation between the distance of losses and \(\lambda^2\). 

\subsection{Peeking Behind the Curtains of Poor Convergence}\label{measurement 2}

Referring to the part of \textit{FL vs. traditional DML} detailed in Sec.~\ref{II}, we identify four key differences between FL and traditional DML. Our objective is to explore which of these factors (or a combination of factors) leads to the failure of the hard-threshold compression. For this purpose, we adopt a full factorial experimental design, an approach that systematically tests all possible combinations of the factors under consideration. We conduct experiments using both the inclusion and exclusion of these four factors, resulting in 16 experiments in Fig.~\ref{motivate fig b}.

\textit{Decaying-$\gamma$ in non-IID scenarios leading to overly aggressive compression during the late training.} As the convex and non-convex ones come up with the same conclusion, we take the convex case as an example in the following content.
In Fig.~\ref{motivate fig b}, we observe a unique phenomenon where the relative compression ratio initially increases and then decreases during the whole training process, occurring under the simultaneous conditions of the decaying-$\gamma$ and non-IID scenarios. In IID settings, the model converges rapidly in the early stage, allowing the hard-threshold compression to transmit a substantial amount of parameters before convergence. Once convergence is achieved, the algorithm automatically inhibits the transmission of extraneous gradients, thereby striking an efficient balance. However, in non-IID scenarios, the model converges more slowly. The decaying-$\gamma$ leads to smaller updates, which, when faced with a fixed threshold, result in increasingly aggressive compression strategies. This leads to a stagnation in model convergence during the late training.
The settings of infrequent communication and partial node participation do not affect the trend of the compression ratio curve, but only alter the relative magnitude of the compression ratio.

\section{The Stepsize-Aware Hard-Threshold Compression in FL} \label{section r-fedht}


We aim to develop a low-cost adaptive hard-threshold compressor less sensitive to decaying-$\gamma$. We denote the threshold at the $t$-th iteration as $\lambda_t$, and the technical challenge is: \textit{How to determine $\lambda_t$ as a function of $\gamma_t$}, \textit{i.e.}, let $\lambda_t^2 = \lambda_0^2 F(\gamma_t)$, how do we determine $F(\gamma_t)$\footnote{The reason for using $\lambda_t^2$ instead of $\lambda_t$ is that the compression error introduced by the hard-threshold compression is linearly related to $\lambda_t^2$ rather than to $\lambda_t$.}? It is difficult to carve $F(\gamma_t)$ out of simple functions because $\lambda_t$ needs to satisfy the following mathematical properties: $\lambda_t$ should initially increase and subsequently diminish to $0$. This can be described as: (1) $\lim_{t->+\infty} F(\gamma_t) = 0$; (2) $\exists 0< t_1\leq t_2 < T$, $\forall t \in (0,t_1)$, $\frac{\mathrm{d} F }{\mathrm{d} t} \geq 0$ and $ \forall t \in (t_2,T)$, $\frac{\mathrm{d} F}{\mathrm{d} t} \leq 0$. The initial increase aligns with the recommendation of employing conservative compression during the initial stages of training \cite{accordion}, and the subsequent decrease aims to slow down the decline of the compression ratio.

In order to reduce the construction space to get $F(\cdot)$, we let $t_1$ = $t_2$ and start with simple functions. Since a single class of simple functions cannot satisfy both properties, we consider a combination of two functions. We choose the inverse proportional function to fulfill the limit of $0$. For the increasing and then decreasing monotonicity, we can select the quadratic function or the logarithmic function. We note that the combination of the inverse proportional function and the logarithmic function, \textit{i.e.}, $F(\gamma_t) = (\gamma_t^\alpha+\gamma_t^{-\alpha})^{-1}$ ($\alpha$ is a constant and $\alpha\geq 1$), can satisfy the mathematical properties while introducing only one hyperparameter.
Additionally, we normalize this function with the geometric mean of $\gamma_t$ and determine $\lambda_{t}^2 = \lambda_0^2 \cdot \frac{\gamma_t^{\alpha} (\gamma_0 \gamma_T)^{\frac{\alpha}{2}}}{\gamma_t^{2\alpha} + (\gamma_0 \gamma_T)^{\alpha}}$, where $\gamma_0$ (as well as $\gamma_T$) is the start (end) value of $\gamma$. 

Based on this, we propose $\gamma$-FedHT, with a time complexity of $\mathcal{O}(1)$ for calculating $\lambda_t$ and $\mathcal{O}(d)$ for compressing gradients. Compared to the compression cost $\mathcal{O}(d)$, the time required to compute $\lambda_t$ is negligible, so \textit{the compression cost of $\gamma$-FedHT is as low as the hard-threshold compressor}. We show the convergence analysis in Sec.~\ref{Theoretical analysis of FedHT}. \textit{We reveal that the convergence rate is fastest when $\alpha=1$}. In other words, there is only one hyperparameter $\lambda_0$ in the $\gamma$-FedHT.   

The pseudo-code of $\gamma$-FedHT is shown in Algo.~\ref{r-FedHT}. $\textbf{C}_{\lambda_{t}}(\textbf{x})$ means compressing the tensor $\textbf{x}$ with the absolute compressor and the compression threshold is $\lambda_{t}$. 

\begin{algorithm}[t]
    \SetAlgoLined 
	\caption{$\gamma$-FedHT}
	\label{r-FedHT}
    \KwIn{number of workers $n$, training weight $p_i$, initial parameters $\textbf{x}_0$, stepsize $\gamma_t$,  absolute compressor $\textbf{C}_{\lambda}(\cdot)$, initial threshold $\lambda_0$, $\alpha$, communication frequency $E$, initial local error $\textbf{e}_0^i = \textbf{0}_d$}
	\KwOut{$\textbf{x}_T$}
	\For{$t = 0, \ldots , T - 1 $}{
             \If{ $t \mod E = 0$ }{
                Server picks nodes $\mathcal{S}_t$ uniformly at random\;
             }
	\tcc{Worker side}
        \For{$i \in \mathcal{S}_t $}{
              \If{ $t \mod E = 0$ }{
                Download $\textbf{x}_t$ from the server\;
             }
             $\textbf{x}_{t+1}^i := \textbf{x}_t^i - \gamma_t\textbf{g}^i(\textbf{x}_t)$, $\textbf{e}_{t+1}^i := \textbf{e}_t^i$\;
             
             
             \If{ $t \mod E = E-1 $ }{
                $\lambda_{t+1} = \lambda_0 \sqrt{\frac{\gamma_{t+1}^{\alpha} (\gamma_0 \gamma_T)^{\alpha/2}}{\gamma_{t+1}^{2\alpha} + (\gamma_0 \gamma_T)^{\alpha}}}$\;
    	    
                $\hat{\Delta}_t^i := \textbf{C}_{\lambda_{t+1}} (\textbf{e}_t^i + \textbf{x}_{t+1}^i - \textbf{x}_{t+1-E}^i)$\;
    	    
                $\textbf{e}_{t+1}^i := \textbf{e}_t^i + \textbf{x}_{t+1}^i - \textbf{x}_{t+1-E}^i - \hat{\Delta}_t^i $\;
    	    
                Upload $\hat{\Delta}_t^i$\;
             }
    	    }
    	    {
    	    \tcc{Server side}
                \eIf{ $t \mod E = E-1$ }{
        	    Gather $\hat{\Delta}_t^i$\ from nodes in $\mathcal{S}_t$\;
        	    
                    $\textbf{x}_{t+1} := \textbf{x}_t -\frac{n}{|\mathcal{S}_t|}\sum_{i\in \mathcal{S}_t}p_i\hat{\Delta}_t^i$\;
        	    
                    Broadcast $\textbf{x}_{t+1}$\;
                 }{
                  $\textbf{x}_{t+1} := \textbf{x}_t$\;
                 }

    	    }
	}
	\textbf{Return} $\textbf{x}_T$\;
\end{algorithm}

\section{Theoretical analysis} \label{Theoretical analysis of FedHT}

\subsection{Regularity Assumptions}
\label{assumptions}

We follow assumptions, which are standard and widely accepted in the theoretical framework of DML \cite{cui2022infocom,stich2022sharper,li2019convergence}. 


\noindent \textbf{Assumption 1} ($L$-smoothness). We assume $L$-smoothness of $f_i, i\in [n]$, that is for all $\textbf{x},\textbf{y} \in \mathbb{R}^d$:
\begin{equation}
\lVert \nabla f(\textbf{y}) - \nabla f(\textbf{x}) \rVert \leq L\lVert \textbf{y} - \textbf{x} \rVert.
\end{equation}

\noindent \textbf{Assumption 2} (Bounded gradient noise). We assume that stochastic gradient oracles $\textbf{g}^i(\textbf{x}): \mathbb{R}^d \rightarrow  \mathbb{R}^d$ are available for each $f_i,i\in [n]$. For simplicity, we only consider the instructive case where the noise is uniformly bounded for all $\textbf{x}, \in \mathbb{R}^d, i \in [n]$:
\begin{equation}
\textbf{g}^i(\textbf{x})=\nabla f_i(\textbf{x})+\boldsymbol{\xi} ^i,
\quad \mathbb{E}_{\boldsymbol{\xi} ^i}\boldsymbol{\xi} ^i = \textbf{0}_d, \quad \mathbb{E}_{\boldsymbol{\xi} ^ i} \lVert \boldsymbol{\xi} ^i \rVert^2 \leq \sigma^2.
\end{equation}

\noindent \textbf{Assumption 3} (Bounded gradient norm). We assume the expected squared norm of stochastic gradients is uniformly bounded:
\begin{equation}
\mathbb{E}_{\boldsymbol{\xi} ^ i} \lVert \textbf{g}^i(\textbf{x}) \rVert^2 \leq G^2,
\end{equation}
where $G$ stands for the upper bound of the gradient norm.

\noindent \textbf{Assumption 4} ($\mu$-strongly convexity). We assume $\mu$-strong convexity of $f_i, i\in [n]$, that is for all  $\textbf{x},\textbf{y} \in \mathbb{R}^d$:
\begin{equation}
f(\textbf{x}) - (\textbf{y}) \geq  \langle \nabla f(\textbf{y}),\textbf{x}-\textbf{y} \rangle + \frac{\mu}{2} \lVert \nabla f(\textbf{y}) - \nabla f(\textbf{x}) \rVert^2.
\end{equation}


For the convex cases 
In convex cases (Theorem~\ref{r-FedHT convex}), we apply Assumptions $1$-$4$ and use $\Gamma_c = f^* - \sum_{i=1}^n p_i f_i^*$ to measure the data heterogeneity. $f^*$ (as well as $f_i^*$) is the optimal value of $f$ ($f_i$) referring to previous works \cite{cui2021slashing,cui2022infocom}. In non-convex cases, we apply Assumptions $1$-$3$ and use $\Gamma_n \leq \mathbb{E}\lVert \nabla f_i(\textbf{x}_t) - \nabla f(\textbf{x}_t)\rVert^2$ to quantize the non-IID degree of nodes referring to the work \cite{linearspeedFL}.

\subsection{Convergence Rate of $\gamma$-FedHT}
\label{convergence rate}


The technical challenge in proving this theorem lies in \textit{integrating gradient compression algorithms with EF into the theoretical framework of FL} \cite{li2019convergence}. 
To tackle this problem, \textit{it is crucial to establish relationship between \(\textbf{x}_{t+1}^i\) and \(\textbf{x}_{t}^i\)}. When aggregation is not performed (\textit{i.e.}, for \(t \mod E \neq E-1\)), we have \(\textbf{x}_{t+1}^i := \textbf{x}_t^i - \gamma_t\textbf{g}^i(\textbf{x}_t)\), consistent with vanilla FedAVG. In aggregation rounds (\textit{i.e.}, for \(t \mod E = E-1\)), we have:
\begin{eqnarray}
    \textbf{x}_{t+1}^i + \sum_{i\in \mathcal{S}_{t+1}}\frac{1}{S}\textbf{e}_{t+1}^i = \sum_{i\in \mathcal{S}_t}\frac{1}{S}[\textbf{x}_{t}^i + \textbf{e}_{t}^i - \gamma_t \textbf{g}^i(\textbf{x}_t^i)], \nonumber
\end{eqnarray}
which can be derived by lines 13 and 19 in Algo.~\ref{r-FedHT}. By the above conversation equation and the virtual sequence method (also known as perturbed iterate analysis), we derive the convergence rate of $\gamma$-FedHT.

\begin{theorem}[$\mu$-strongly convex Convergence rate of $\gamma$-FedHT]
\label{r-FedHT convex}
Let $f: \mathbb{R}^d \rightarrow \mathbb{R}$ be $L$-smooth and $\mu$-convex. Choose $\kappa = \frac{L}{\mu}$, $b=\max \{12\frac{L}{\mu},E\}-1$ and the stepsize $\gamma_t=\frac{3}{\mu(b+t)}$. Then $\gamma$-FedHT satisfies
\begin{eqnarray}
    && \mathbb{E}[f(\textbf{x}_T)] - f^* \leq\frac{\kappa}{t+b} \{\frac{9}{\mu}[(1+\frac{\gamma_0 \mu}{2})B  \nonumber\\
   && +(\frac{1}{2}+ \frac{2(\gamma_0/\gamma_T)^{\alpha/2}}{\mu\gamma_0})D] + \frac{\mu (1+b)}{2} \Delta_1 \},
\end{eqnarray}
where $B=\sum_{i=1}^n p_i^2\sigma^2 + 6L\Gamma + 8(E-1)^2 G^2 + \frac{4}{S}E^2G^2 $ and  $D=4d \lambda_0^2$.
\end{theorem}

\begin{remark}
The convergence rate of $\gamma$-FedHT is $\mathcal{O}(\frac{1}{T})$ under the $\mu$-strongly convex cases, same as vanilla FedAVG \cite{li2019convergence}.
\end{remark}

\begin{remark}
The term $(\frac{1}{2}+ \frac{2(\gamma_0/\gamma_T)^{\alpha/2}}{\mu\gamma_0})D$  is the bound of the compression error, which is linearly correlated with $\lambda_0^2$. When $\lambda_0 = 0$, $\gamma$-FedHT degrades to vanilla FedAVG.
\end{remark}

\begin{remark}
\label{remark 1}
The larger the $\alpha$, the more iterations are needed for the convergence. So \textit{we take $\alpha=1$ by default} for $\alpha \geq 1$ in the $\mu$-strongly convex cases.
\end{remark}

\begin{theorem}[Non-Convex Convergence rate of $\gamma$-FedHT]
\label{r-FedHT non-convex}
Let $f: \mathbb{R}^d \rightarrow \mathbb{R}$ be $L$-smooth. Choose $c > 0$, and $\gamma_t \leq \frac{1}{8LE}$. $\gamma_t$ satisfies $\gamma_t E L \leq \frac{2S}{S-1}$ and $30nE^2 \gamma_t^2 L^2 \sum_{i=1}^n p_i^2 + \frac{2L\gamma_t}{S}(90E^3 L^2 \gamma_t^2 +3E) < 1$. The convergence rate of $\gamma$-FedHT satisfies 
\begin{eqnarray}
    &&\underset{t \in [T]}{\min} \mathbb{E} \lVert \nabla f (\textbf{x}_t) \rVert^2 \leq \frac{f_0 - f^*}{c\gamma_{T-1}TE} + (LE^2\sigma^2 +  \frac{3E^2 L \Gamma_n}{S} )\nonumber \\
    &&\frac{\sum_{t=0}^{T-1}\gamma_{t}^2}{c\gamma_{T-1}TE} + \frac{2Ld\lambda_0^2}{\gamma_0}(\frac{\gamma_0}{\gamma_{T-1}})^{\frac{\alpha}{2}}\frac{\sum_{t=0}^{T-1}\gamma_{t}^3}{c\gamma_{T-1}TE}+
    \frac{\sigma^2+6E \Gamma_n}{c\gamma_{T-1}TE} \nonumber\\
   &&[\frac{5nE^2 L^2 \sum_{i=1}^n p_i^2 \sum_{t=0}^{T-1}\gamma_{t}^3 }{2} + \frac{15E^3L^3\sum_{t=0}^{T-1}\gamma_{t}^4}{S}]. \nonumber
\end{eqnarray}

\begin{remark}
The convergence rate of $\gamma$-FedHT is $\mathcal{O}(\frac{1}{\sqrt{T}})$ under non-convex cases, also same as FedAVG \cite{linearspeedFL}.
\end{remark}

\begin{remark}
The term $\frac{LD}{2\gamma_0}(\frac{\gamma_0}{\gamma_{T-1}})^{\alpha/2}\frac{\sum_{t=0}^{T-1}\gamma_{t}^3}{c\gamma_{T-1}TE}$ represents the compression error and is bounded by $\lambda_0^2$. When $\lambda_0 = 0$, the convergence of $\gamma$-FedHT degrades to that of FedAVG.
\end{remark}

\begin{remark}
\label{remark 2}
Due to $\frac{\gamma_0}{\gamma_{T-1}}>1$, $\gamma$-FedHT converges the fastest when $\alpha=1$.\textit{ We take $\alpha=1$ by default in the non-convex cases too}. 
\end{remark}

\end{theorem}

\subsection{Proof of Theorem~\ref{r-FedHT convex}}
\label{proof convex}

Let $\mathcal{I}_E$ be the set of communication iterations. We have $\mathcal{I}_E = \{i E| i = 1,2,\dots \}$. Here we introduce a variable $\textbf{v}_{t+1}^i$ to represent the result of local SGD from $\textbf{x}_{t+1}^i$. Then the training of Algo. \ref{r-FedHT} can be described as 
\begin{equation}
\label{v->x}
    \textbf{v}_{t+1}^i = \textbf{x}_t^i - \gamma_t \textbf{g}^i(\textbf{x}_t^i),
\end{equation}

\begin{small}
\begin{equation}
\label{x->v}
    \textbf{x}_{t+1}^i = \left\{
             \begin{array}{ll}
             \! \textbf{v}_{t+1}^i & \! \textrm{if } t+1 \notin \mathcal{I}_E,\\
             \! \sum_{i\in \mathcal{S}_t} \frac{1}{S} (\textbf{v}_{t+1}^i + \textbf{e}_{t}^i )  - \sum_{i\in \mathcal{S}_{t+1}}\frac{1}{S}\textbf{e}_{t+1}^i & \! \textrm{if } t+1 \in  \mathcal{I}_E.
             \end{array}
\right. 
\end{equation}
\end{small}


Motivated by previous works, we introduce two virtual sequences $\Bar{\textbf{v}}_{t}=\sum_{i=1}^n p_i \textbf{v}_{t}^i$ and $\Bar{\textbf{x}}_{t}=\sum_{i=1}^n p_i \textbf{x}_{t}^i$. 
We use the notations  $\textbf{E}_{t} = \mathbb{E}\lVert \frac{1}{S}\sum_{i\in \mathcal{S}_t} \textbf{e}_{t+1}^i \rVert^2$ and $\Delta_t= \mathbb{E}\lVert \Bar{\textbf{x}}_t - \textbf{x}^* \rVert^2$.



According to Lemma 1-5 from the work \cite{li2019convergence}, we have:

\begin{lemma}
\label{v->delta}
Following the Assumption 1-4. If $\gamma_t \leq \frac{1}{4L}$, $\gamma_t$ is non-increasing and $\gamma_t \leq 2 \gamma_{t+E}$ for all $t \geq 0$, we have
    \begin{equation}
         \mathbb{E}\lVert \Bar{\textbf{v}}_{t+1} - \textbf{x}^* \rVert^2 \leq (1-\gamma_t \mu) \Delta_{t} + \gamma_t^2 B,
    \end{equation}
where $B=\sum_{i=1}^n p_i^2\sigma^2 + 6L\Gamma + 8(E-1)^2 G^2 + \frac{4}{S}E^2G^2$.

\end{lemma}

Before the proof, we compare Eq.~\ref{v->x} and Eq.~\ref{x->v} with the Sec.~A-3 in \cite{li2019convergence}. We can find that $\Bar{\textbf{v}}_{t+1} = \Bar{\textbf{x}}_{t} - \gamma_t \textbf{g}_t$ no matter $t+1 \notin \mathcal{I}_E$ or $t+1 \in \mathcal{I}_E$.  Then we categorize and discuss two cases (\textit{i.e.}, $t+1 \notin \mathcal{I}_E$ and $t+1 \in \mathcal{I}_E$) because Eq. \ref{x->v} is different from \cite{li2019convergence}. 

\noindent$\bullet$   \textit{$t+1 \notin \mathcal{I}_E$:} Due to  $\Bar{\textbf{v}}_{t+1} = \Bar{\textbf{x}}_{t+1}$ and Lemma~\ref{v->delta}, we have
\begin{equation}
\label{D->Dold}
    \Delta_{t+1} \leq (1-\gamma_t \mu) \Delta_{t} + \gamma_t^2 B.
\end{equation}

\noindent$\bullet$   \textit{$ t+1 \in  \mathcal{I}_E $:} Due to $\Bar{\textbf{x}}_{t+1} = \Bar{\textbf{v}}_{t+1} - \frac{1}{S}\sum_{i\in \mathcal{S}_t} \textbf{e}_{t+1}^i$, we have
\begin{eqnarray}
\label{x->delta}
   \mathbb{E}\lVert \Bar{\textbf{x}}_{t+1} - \textbf{x}^* \rVert^2 &\leq& (1-\frac{\gamma_t \mu }{2}) \Delta_t + (1+\frac{\gamma_t \mu }{2}) \gamma_t^2 B  \nonumber \\&& + 2(1+\frac{2}{\gamma_t \mu})(\textbf{E}_{t+1} + \textbf{E}_{t}).
\end{eqnarray}

The inequality is followed by Jensen inequality and Lemma~\ref{v->delta}. 

We next focus on $\textbf{E}_t$ and have
\begin{eqnarray}
\label{E}
   \textbf{E}_{t}  \leq \frac{1}{S} \sum_{i\in \mathcal{S}_t} \lVert \textbf{e}_t^i \rVert^2 \leq  \gamma_t^2 d \lambda_t^2 \leq \gamma_t^2 d \lambda_0^2 F(\gamma_{t-1}), 
\end{eqnarray}
where the first inequality is due to 
$\lVert \sum_{i=1}^k a_i \rVert^2 \leq k \sum_{i=1}^k \lVert a_i\rVert^2$. The second inequality follows the property of the absolute compressor.

We combine Eq. \ref{x->delta} and Eq. \ref{E} and have
\begin{eqnarray}
\label{x->delta+E}
    &&\mathbb{E}\lVert \Bar{\textbf{x}}_{t+1} - \textbf{x}^* \rVert^2 \leq (1-\frac{\gamma_t \mu }{2}) \Delta_t + (1+\frac{\gamma_t \mu }{2}) \gamma_t^2 B \nonumber\\
    & &+ (1+\frac{2}{\gamma_t \mu})\gamma_{t+1}^2 d \lambda_0^2 (F(\gamma_{t})+F(\gamma_{t+1})) \nonumber\\
    &&\leq (1-\frac{\gamma_t \mu }{2}) \Delta_t + (1+\frac{\gamma_t \mu }{2}) \gamma_t^2 B + [(1+\frac{2}{\gamma_t \mu}) \gamma_{t}^2 F(\gamma_{t}) \nonumber\\
    &&+ (1+\frac{2}{\gamma_{t+1} \mu}) \gamma_{t+1}^2 F(\gamma_{t+1}) ]D,
\end{eqnarray}
where $B=\sum_{i=1}^n p_i^2\sigma^2 + 6L\Gamma + 8(E-1)^2 G^2 + \frac{4}{S}E^2G^2 $  and $D=2d \lambda_0^2$.

The second inequality is due to the non-increasing $\gamma_t$. 

Since the RHS of Eq. \ref{D->Dold} is smaller than the RHS of Eq. \ref{x->delta+E}, the RHS of  Eq. \ref{x->delta+E} applies for all $t\leq 0$. We have
\begin{small}
\begin{align}
    \Delta_{t+1} &\leq
    (1-\frac{\gamma_t \mu }{2}) \Delta_t + (1+\frac{\gamma_t \mu }{2}) \gamma_t^2 B \nonumber \\
    &\quad + \left[(1+\frac{2}{\gamma_t \mu}) \gamma_{t}^2 F(\gamma_{t}) + (1+\frac{2}{\gamma_{t+1} \mu}) \gamma_{t+1}^2 F(\gamma_{t+1}) \right]D.
\end{align}
\end{small}

For a decaying learning rate, $\gamma_t = \frac{\beta}{t+b}$ for some $\beta > \frac{2}{\mu}$ and $b > 0$ such that $\gamma_0 \leq \frac{1}{4L}$ and $\gamma_t \leq 2 \gamma_{t+E}$. We utilize the mathematical induction to prove $\Delta_t \leq \frac{v}{b+t}$, where $v = \max \{\frac{2\beta^2 }{\beta \mu - 2}[(1+\frac{\gamma_0 \mu}{2})B + (\frac{1}{2} + \frac{2\gamma_{0}^{\alpha-1}}{\mu (\gamma_0 \gamma_T)^{\alpha/2}}) 4d\lambda_0^2 ],(1+b)\Delta_1\}$.

When $t = 1$,  $\Delta_t \leq \frac{v}{b+t}$ clearly holds.

When $t > 1$, it follows that
\begin{small}
\begin{equation}
\begin{aligned}
\Delta_{t+1} 
   &\leq (1-\frac{\gamma_t \mu }{2})\frac{v}{t+b} + (1+\frac{\gamma_0\mu }{2}) \gamma_t^2 B \\
   &\quad + \left[(1+\frac{2}{\gamma_t \mu}) \gamma_{t}^2 F(\gamma_{t}) + (1+\frac{2}{\gamma_{t+1} \mu}) \gamma_{t+1}^2 F(\gamma_{t+1}) \right] D \\
   &\leq \frac{v}{t+b+1} + \left[(1+\frac{\gamma_0\mu }{2}) \gamma_t^2 B - \frac{\beta \mu - 2}{2(t+b)^2} v \right] \\
   &\quad + \underbrace{\left[(1+\frac{2}{\gamma_t \mu}) \gamma_{t}^2 F(\gamma_{t}) + (1+\frac{2}{\gamma_{t+1} \mu}) \gamma_{t+1}^2 F(\gamma_{t+1}) \right] D}_{A_1}.
\end{aligned}
\label{induct_1}
\end{equation}
\end{small}

The first inequality holds by the inductive conclusion $\Delta_t \leq \frac{v}{b+t}$ and $\gamma_t \leq \gamma_0$.

We next aim to bound $A_1$. According to $\gamma$-FedHT, we have 
\begin{small}
\begin{equation}
\begin{aligned}
    A_1 
    &\leq D( \frac{\gamma_t^2}{2} + \frac{2\gamma_t^{\alpha+1}}{\mu (\gamma_0 \gamma_T)^{\alpha/2}}) + D( \frac{\gamma_{t+1}^2}{2} + \frac{2\gamma_{t+1}^{\alpha+1}}{\mu \sqrt{\gamma_0 \gamma_T}^{\alpha}}) \\
    &\leq 2D( \frac{\gamma_t^2}{2} +\frac{2\gamma_{t}^2 \gamma_{0}^{\alpha-1}}{\mu (\gamma_0 \gamma_T)^{\alpha/2}}).
\end{aligned}
\label{A_1}
\end{equation}
\end{small}

The first inequality is due to the arithmetic-geometric mean inequality (the first part) and $\gamma_t > 0$ (the second part). The second inequality is due to the decaying-$\gamma$.

Combining Eq. \ref{A_1} and Eq. \ref{induct_1}, we have 
\begin{eqnarray}
   \Delta_{t+1}
   & \leq & \frac{v}{t+b+1} + \gamma_t^2 \{\frac{2\beta^2 }{\beta \mu - 2}[(1+\frac{\gamma_0 \mu}{2})B \nonumber\\
   &&+ (\frac{1}{2} + \frac{2\gamma_{0}^{\alpha-1}}{\mu (\gamma_0 \gamma_T)^{\alpha/2}}) D' ]- v \} 
   \leq \frac{v}{t+b+1}, \nonumber 
\end{eqnarray}
which completes the proof of $\Delta \leq \frac{v}{t+b}$ and $D' = 2D = 4d\lambda_0^2$.

According to the $L$-smoothness of $f(\cdot)$, 
\begin{eqnarray}
    \mathbb{E}[f(\textbf{x}_T)] - f^* \leq \frac{L}{2} \Delta_t \leq \frac{L}{2} \frac{v}{t+b}.\nonumber
\end{eqnarray}

We let $\beta = \frac{3}{\mu}$, $b=\max \{12\kappa,E\}-1$ ($\kappa = \frac{L}{\mu}$) and have  $v \leq \frac{18}{\mu^2}[(1+\frac{\gamma_0 \mu}{2})B + (\frac{1}{2} + \frac{2\gamma_{0}^{\alpha-1}}{\mu (\gamma_0 \gamma_T)^{\alpha/2}}) D' ] + (1+b) \Delta_1$.

\subsection{Proof of Theorem~\ref{r-FedHT non-convex}}
\label{proof non-convex}

We let $t'=\lfloor \frac{t}{E} \rfloor$, $\Delta_{t'}^i = \sum_{j=0}^{E-1} \nabla f_i (\textbf{x}_{t'+j}^i)$. $t'$ represents the communication iteration, and $\Delta_{t'}^i $ represents the gradients accumulated between the $t'$-th and the $t'+1$-th global iterations at node $i$.

We define a virtual sequence:
\begin{equation}
   \Bar{\textbf{x}}_0 = \textbf{x}_0, \quad \Bar{\textbf{x}}_{t'+1} := \Bar{\textbf{x}}_{t'}-\frac{\gamma_{t'} }{S}\sum_{i \in \mathcal{S}_t} \Delta_{t'}^i \nonumber.
\end{equation}

The error term that represents the deviation between the virtual sequence and the actual sequence is
\begin{equation}
 \Bar{\textbf{x}}_{t'} - \textbf{x}_{t'} = \frac{\gamma_{t'}}{S} \sum_{i=1}^n \textbf{e}_{t'}^i \nonumber.
\end{equation}

Given $L$-smoothness of $f$, we have 
\begin{small}
\begin{equation}
\begin{aligned}
    \label{x_t+1 -> x_t}
    &\mathbb{E} f(\Bar{\textbf{x}}_{t'+1}) 
    \leq   f(\Bar{\textbf{x}}_{t'}) 
    - <\nabla f(\textbf{x}_{t'}), \frac{\gamma_{t'} }{S}\sum_{i \in \mathcal{S}_t} \Delta_{t'}^i >  
    \\
    & +  \underbrace{<\nabla f(\textbf{x}_{t'})-\nabla f(\Bar{\textbf{x}}_{t'}), \frac{\gamma_{t'} }{S}\sum_{i \in \mathcal{S}_t} \Delta_{t'}^i>}_{A_2} + \frac{L}{2} \mathbb{E} \lVert \frac{\gamma_{t'}}{S} \sum_{i=1}^n \Delta_{t'}^i \rVert^2. 
\end{aligned} 
\end{equation}
\end{small}

We next aim to bound $A_2$, where
\begin{eqnarray}
    \label{A_2}
    &&<\nabla f(\textbf{x}_{t'})-\nabla f(\Bar{\textbf{x}}_{t'}), \frac{\gamma_{t'} }{S}\sum_{i \in \mathcal{S}_t} \Delta_{t'}^i> \nonumber \\
    &&
    \leq 
    \frac{1}{2L} \mathbb{E} \lVert f(\textbf{x}_{t'})-\nabla f(\Bar{\textbf{x}}_{t'}) \rVert^2 
    + \frac{L}{2}\mathbb{E} \lVert \frac{\gamma_{t'}}{S} \sum_{i=1}^n \Delta_{t'}^i \rVert^2 \nonumber \\
    &&\leq \frac{L}{2} \textbf{E}_{t'} + \frac{L}{2}\mathbb{E} \lVert \frac{\gamma_{t'}}{S} \sum_{i=1}^n \Delta_{t'}^i \rVert^2.
\end{eqnarray}

The first inequality is followed by Jensen inequality, and the last inequality is held by $L$-smooth functions. 

According to the Appendix B in the work \cite{cui2022infocom},  Eq.~\ref{E} and~\ref{A_2}, we can convert Eq.~\ref{x_t+1 -> x_t} into:
\begin{small}
\begin{eqnarray}
    &&\mathbb{E} f(\Bar{\textbf{x}}_{t'+1})
    \leq  f(\Bar{\textbf{x}}_{t'}) - \gamma_{t'} E\lVert \nabla f (\textbf{x}_{t'})\rVert^2 \nonumber\\
         &&[\frac{1}{2} - 15nE^2\gamma_{t'}^2L^2 \sum_{i=1}^n p_i^2 - \frac{L\gamma_{t'}}{S} (90E^3L^2 \gamma_{t'}^2 + 3E) ] \nonumber\\
         &&+ (\frac{5nE^2\gamma_{t'}^3 L^2 \sum_{i=1}^n p_i^2 }{2} + \frac{15E^3L^3\gamma_{t'}^4}{S}) (\sigma^2 + 6E\Gamma_n)  \nonumber\\
         && + LE^2\gamma_{t'}^2\sigma^2 + \frac{3E^2 L  \gamma_{t'}^2 \Gamma_n}{S} + \frac{LD}{2\gamma_0}   (\frac{\gamma_0}{\gamma_{T-1}})^{\alpha/2} \gamma_{t'}^3\nonumber\\
         &&+ (\frac{L\gamma_{t'}^2(S-1)}{S} - \frac{\gamma_{t'}}{2E}) \mathbb{E} \lVert \sum_{i=1}^n p_i \sum_{j=0}^{E-1} \nabla f_i (\textbf{x}_{t'+j}^i) \rVert ^2   \nonumber \\
    &&\leq f(\Bar{\textbf{x}}_{t'}) - c \gamma_{t'} E\lVert \nabla f (\textbf{x}_{t'})\rVert^2 \nonumber \\
    &&+(\frac{5nE^2\gamma_{t'}^3 L^2 \sum_{i=1}^n p_i^2 }{2} + \frac{15E^3L^3\gamma_{t'}^4}{S}) (\sigma^2 + 6E\Gamma_n) \nonumber\\
    &&  + LE^2\gamma_{t'}^2\sigma^2+ \frac{3E^2 L  \gamma_{t'}^2 \Gamma_n}{S} + \frac{LD}{2\gamma_0}   (\frac{\gamma_0}{\gamma_{T-1}})^{\alpha/2} \gamma_{t'}^3. \nonumber
\end{eqnarray}
\end{small}

The last inequality follows from $\frac{L\gamma_{t'}^2(S-1)}{S} - \frac{\gamma_{t'}}{2E} \leq 0 $ if $\gamma_{t'}EL \leq \frac{S}{2}$ and $\frac{1}{2} - 15nE^2\gamma_{t'}^2L^2 \sum_{i=1}^n p_i^2 - \frac{L\gamma_{t'}}{S} (90E^3L^2 \gamma_{t'}^2 + 3E) > c > 0$ if $15nE^2\gamma_{t'}^2L^2 \sum_{i=1}^n p_i^2 + \frac{L\gamma_{t'}}{S} (90E^3L^2 \gamma_{t'}^2 + 3E) < \frac{1}{2}$. We complete the proof.

\section{Evaluation Experiments} \label{V}


\begin{figure}[!t]
\centering
\subfigure{
\hspace{-0.5em}\includegraphics[width=0.99\linewidth]{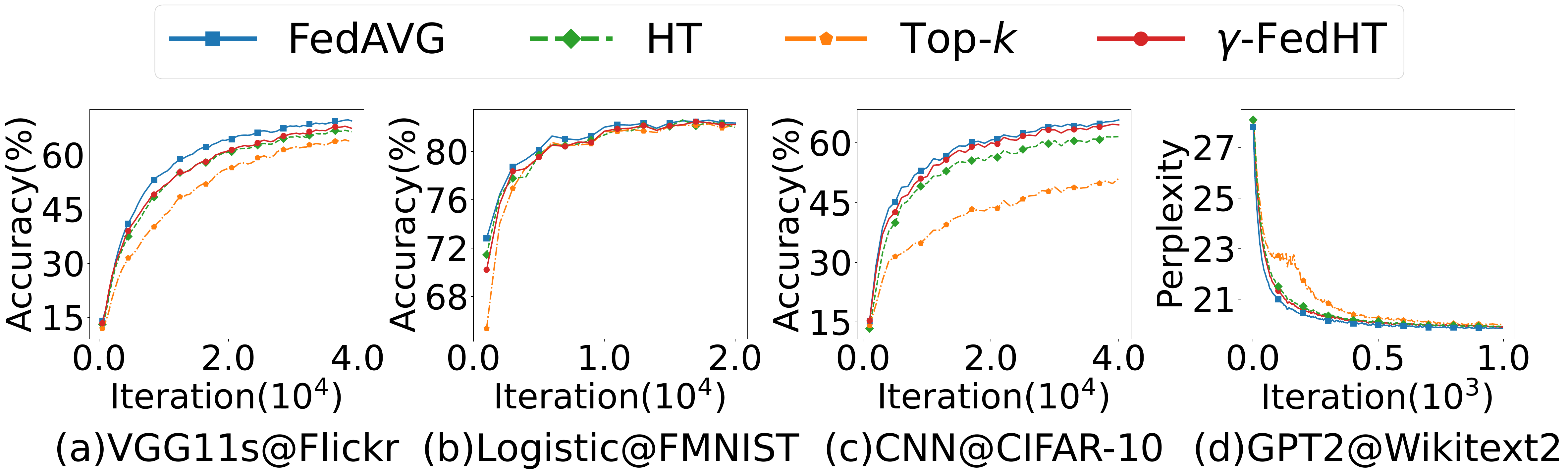}}
\caption{Training curves (Accuracy vs. Iterations). The artificially non-IID partition strategy is $\#C=2$. On all benchmarks, $\gamma$-FedHT outperforms hard-threshold compression (HT) and Top-$k$.}
\label{fig:experiment}
\vspace{-1em}
\end{figure}

\begin{figure}[t]
\centering
\hspace{-0.3em}\includegraphics[scale=0.148]{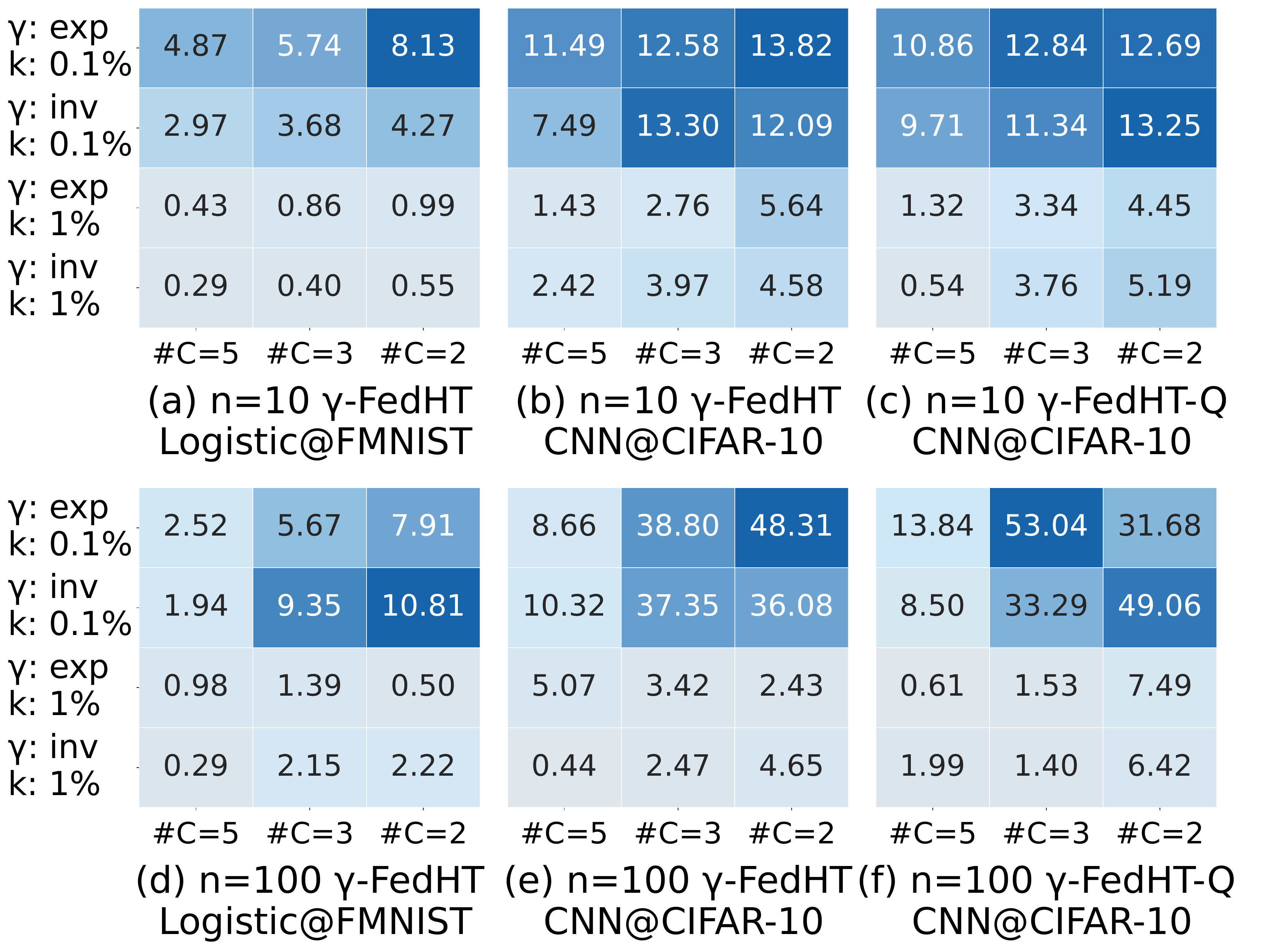}
\caption{Heatmaps of the accuracy difference ($\%$) for different types of stepsizes (denoted as $\gamma$), compression ratios ($k$), worker size ($n$) on different tasks. The accuracy difference in (a, b, d, e) is the final accuracy of $\gamma$-FedHT minus that of Top-$k$, and the difference in (c, f) is $\gamma$-FedHT-Q minus STC. In all combinations, $\gamma$-FedHT (as well as $\gamma$-FedHT-Q) is superior to Top-$k$ (STC).}
\label{fig:experiment2}
\vspace{-1em}
\end{figure}


\begin{table*}[!th]
\captionsetup{justification=raggedright, singlelinecheck=false}
\caption{Summary of the experiment settings used in this work.}
\label{table:setting}
\centering
\begin{threeparttable}
\begin{tabular}{cccccccccc}
   \toprule
   Task & Model & Model parameters & Dataset & Non-IID type & Loss convexity & Batch size & $n$ & Metric &  Iterations  \\
   \midrule
   \multirow{3}{*}{CV} & 
    VGG11s \cite{sbc} &$865,482$  & Flickr \cite{niid2020icml} &  Real-world & Non-convex loss & $8$ & $15$ & \multirow{3}{*}{ Accuracy} & $40,000$ \\
    & Logistic \cite{dagc} 
   & $10,250$ & FMNIST \cite{fmnist}  & Artificially 
   & Non-convex loss & $50$ & $10$ \& $100$ 
   &  & $20,000$ \\  
   & CNN \cite{stc}
   &  $235,690$ & CIFAR-$10$
   \cite{cifar10}  &  Artificially 
   & Convex loss & $8$ & $10$ \& $100$ &  &  $40,000$ \\
   \midrule
   NLP &   GPT2 \cite{gpt2} 
   &   $124,000,000$ & Wikitext2 \cite{wikitext} & Artificially 
   & Non-convex loss & $1$ & $10$ & Perplexity  &  $1,000$\\

   \bottomrule
   
\end{tabular}

\end{threeparttable}
\label{benchmark}
\end{table*}


\subsection{Experimental Settings}\label{V-A}


\noindent \textbf{Experiment tasks:}
We conduct experiments on the four tasks and the detailed setting is shown in Table~\ref{table:setting}. 

\noindent \textbf{Non-IID partition strategy:} For the artificially non-IID partition, we adopt $3$ distinct non-IID partition strategies, namely $\#C=2$, $\#C=3$, $\#C=5$. For Flickr, we divide workers according to the subcontinent they belong to, with 15 clients in total.

\noindent \textbf{Baselines:} We compare (1) $\gamma$-FedHT with Top-$k$, the hard-threshold compressor with fixed-$\lambda$ (denoted as HT), and vanilla FedAVG; (2) $\gamma$-FedHT-Q ($\gamma$-FedHT followed by the quantizer used in STC) with STC. Top-$k$ is the SOTA sparsification gradient compressor in FL and serves as a key component of nowadays hybrid gradient compressors \cite{cui2021slashing,stc, FedZIP}. HT is the SOTA sparsifier in traditional DML \cite{hardthreshold}. FedAVG without compression is used as the benchmark for evaluation. STC is the SOTA hybrid compressor in FL \cite{stc}.  \textit{We take $\alpha=1$ in $\gamma$-FedHT due to Remark~\ref{remark 1} and Remark~\ref{remark 2}}.

\noindent \textbf{Hyperparameters:} 
For each communication round, we randomly select half of the clients to participate. We configure the communication frequency $E=5$, the inverse-proportional decay stepsize $\gamma_t=\frac{100}{t+1000}$ and exponential decay stepsize $\gamma_t=0.1\times 0.999^{t/E}$ for $n=10$, $15$. For $n=100$, we reduce $\gamma_t$ by a multiple of $10$.

\begin{table}[t]
\centering
\captionsetup{justification=raggedright, singlelinecheck=false}
\caption{Values of compression-related hyperparameters.}
\label{table:appendix}
\scalebox{1}{
\resizebox{0.98\linewidth}{!}{
\begin{tabular}{ccccc}
\toprule
Model@Dataset & Top-$k$ $k$ & HT $\lambda$ & \makecell[c]{$\gamma$-FedHT $\lambda_0$ \\(inv $\gamma$)} & \makecell[c]{$\gamma$-FedHT $\lambda_0$ \\(exp $\gamma$)} \\
\midrule
VGG11s@Flickr & 0.1\% & $1.70 \times 10^{-2}$ & $3.35 \times 10^{-2}$ & $6.28 \times 10^{-2}$ \\
Logistic@FMNIST & 1\% & $4.94 \times 10^{-2}$ & $8.70 \times 10^{-2}$ & $9.41 \times 10^{-2}$ \\
CNN@CIFAR-10 & 0.1\% & $3.26 \times 10^{-2}$ & $6.42 \times 10^{-2}$ & $1.21 \times 10^{-1}$ \\
GPT2@Wikitext2 & 0.1\% & $1.42 \times 10^{-3}$ & $2.29 \times 10^{-3}$ & $9.02 \times 10^{-3}$ \\
\bottomrule
\end{tabular}
}}
\end{table}

\subsection{Comparison of Model Accuracy}\label{V-B}

Our experimental results show that the training results of $\gamma$-FedHT outperform HT and Top-$k$ on all tasks. The compression-related parameters are shown in Table~\ref{table:appendix}. We calculate $\lambda$ and $\lambda_0$ referring to Appendix~D of the work\footnote{We use $\lambda=\frac{1}{2\sqrt{dk}}$ and $\int_{0}^T \frac{1}{\lambda^2}dt = \int_{0}^T \frac{1}{\lambda_t^2}dt$ to work out $\lambda$, $\lambda_0$ respectively.} \cite{hardthreshold}.

In Fig.~\ref{fig:experiment}, we find that $\gamma$-FedHT always converges better than other sparsifiers. Let us take CNN@CIFAR-$10$  as an example. To converge to  $50\%$, $55\%$, $60\%$ accuracy, $\gamma$-FedHT is faster than HT by about $3.75\%$, $4.25\%$, $8.15\%$ iterations, and the accuracy of Top-$k$ cannot reach $60\%$. Vanilla HT and Top-$k$ introduce severe accuracy degradation in this case, but $\gamma$-FedHT does not. 

In Fig.~\ref{fig:experiment2}, we conduct a sensitivity analysis for both Logistic@FMNIST and CNN@CIFAR-$10$ on five factors. Below is a detailed discussion of these factors.

\noindent \textbf{Compression ratio and non-IID partition strategy:} The more aggressive the compression ($k$ from $1\%$ to $0.1\%$) and the more severe the non-IID problem (from $\#C=5$ to $\#C=2$), the larger the accuracy difference is. This shows that \textit{$\gamma$-FedHT greatly alleviates accuracy degradation when faced with extremely aggressive compression and severe non-IID problems}.

\noindent \textbf{Worker size:} 
The variance of the accuracy differences at $n=100$ is larger than that at $n=10$. This is interesting because it illustrates that a larger worker size can have two effects at the same time. Firstly, when confronted with accuracy degradation, a larger $n$ tends to enlarge this degradation. This can be seen from outliers ($>30\%$) in (e, f), which indicate that Top-$k$ does not converge. In contrast, \textit{$\gamma$-FedHT is surprisingly robust to large-scale FL training}. Secondly, a larger $n$ also accelerates the model convergence (by training more data in one iteration), thus reducing the accuracy differences. 

\noindent \textbf{Decaying type of the stepsize and whether to bring the quantizer or not:} Carrying a quantizer essentially makes the compression more aggressive, whereas $\gamma$-FedHT is robust to aggressive compression environments, so the difference is further enlarged in $\gamma$-FedHT-Q in (c, f). The decaying type of $\gamma$ does not affect the excellent compression-accuracy trade-off of $\gamma$-FedHT.

\begin{table}[!t]
\centering
\caption{Accuracy and communication traffic of different gradient compression algorithms under different non-IID partition strategies. The results show that $\gamma$-FedHT outperforms other sparsifiers under both non-convex and convex cases especially when the communication is restricted and the non-IID problem is extremely severe.}
\label{table:CommW/oQ}
\vspace{0.5em}
\scalebox{0.98}{
\resizebox{\linewidth}{!}{
\begin{tabular}{cccccc}
\toprule
\makecell[c]{Model\\@Dataset}& \makecell[c]{Non-IID\\Partition\\Strategy} & Method & Accuracy & \makecell[c]{Comm. \\Traffic} & \makecell[c]{Comm. Traffic \\Reduced to}\\
\midrule
\multirow{12}*{\makecell[c]{Logistic\\@FMNIST}} & \multirow{5}*{$\#C=2$} & Top-($k_{\mathrm{mean}}$) & $81.97\% $ & $3.44$MB & $2.20\% $ \\
\cline{3-6}
& & HT & $81.99\% $ & $3.78$MB & $2.42\% $ \\
\cline{3-6}
& & \textbf{$\gamma$-FedHT} & $\mathbf{82.23}\% $ & $\mathbf{3.44}$MB & $\mathbf{2.20}\% $ \\
\cline{3-6}
& & FedAVG & $82.34\% $ & 156.40MB & $100\% $ \\
\cline{2-6}
& \multirow{4}*{$\#C=3$} & Top-($k_{\mathrm{mean}}$) & $82.84\% $ & $3.19$MB & $2.04\% $ \\
\cline{3-6}
& & HT & $82.82\% $ & $3.50$MB & $2.24\% $ \\
\cline{3-6}
& & \textbf{$\gamma$-FedHT} & $\mathbf{83.05}\% $ & $\mathbf{3.19}$MB & $\mathbf{2.04}\% $ \\
\cline{3-6}
& & FedAVG & $83.11\% $ & $156.4$MB & $100\% $ \\
\cline{2-6}
& \multirow{4}*{$\#C=5$} & \textbf{Top-($k_{\mathrm{mean}}$)} & $\mathbf{83.56}\% $ & $\mathbf{2.56}$MB & $\mathbf{1.64}\% $\\
\cline{3-6}
& & HT & $83.43\% $ & $2.82$MB & $1.80\% $ \\
\cline{3-6}
& & $\gamma$-FedHT & $83.51\% $ & $2.56$MB & $1.64\% $ \\
\cline{3-6}
& & FedAVG & $83.57\% $ & $156.40$ MB & $100\% $ \\
\midrule


\multirow{12}*{\makecell[c]{CNN\\@CIFAR-$10$}} & \multirow{4}*{$\#C=2$} & Top-($k_{\mathrm{mean}}$) & $57.57\% $ & $23.74$MB & $0.33\% $ \\
\cline{3-6}
& & HT & $61.56\% $ & $21.58$MB & $0.30\% $ \\
\cline{3-6}
& & \textbf{$\gamma$-FedHT} & $\mathbf{64.51}\% $ & $\mathbf{23.74}$MB & $\mathbf{0.33}\% $ \\
\cline{3-6}
& & FedAVG & $65.75\% $ & $7192.69$MB & $100\% $ \\
\cline{2-6}
& \multirow{4}*{$\#C=3$} & Top-($k_{\mathrm{mean}}$) & $64.88\% $ & $25.89$MB & $0.36\% $ \\
\cline{3-6}
& & HT & $71.12\% $ & $30.21$MB & $0.42\% $ \\
\cline{3-6}
& & \textbf{$\gamma$-FedHT} & $\mathbf{72.30}\% $ & $\mathbf{25.89}$MB & $\mathbf{0.36}\% $ \\
\cline{3-6}
& & FedAVG & $72.42\% $ & $7192.69$MB & $100\% $ \\
\cline{2-6}
& \multirow{4}*{$\#C=5$} & Top-($k_{\mathrm{mean}}$) & $70.33\% $ & $20.14$MB & $0.28\% $ \\
\cline{3-6}
& & HT & $73.35\% $ & $23.74$MB & $0.33\% $ \\
\cline{3-6}
& & \textbf{$\gamma$-FedHT} & $\mathbf{74.58}\% $ & $\mathbf{20.14}$MB & $\mathbf{0.28}\% $ \\
\cline{3-6}
& & FedAVG & $75.36\% $ & $7192.69$MB & $100\% $ \\
\bottomrule
\end{tabular}
}
}
\vspace{-1em}
\end{table}

\begin{table}[!t]
    \centering
    \caption{Accuracy and communication traffic of STC and $\gamma$-FedHT-Q under different non-IID partition strategies. The results show that $\gamma$-FedHT-Q outperforms STC under both non-convex and convex cases.}
    \label{table:CommWQ}
    \vspace{0.5em}
    \scalebox{0.98}{
        \resizebox{\linewidth}{!}{
            \begin{tabular}{cccccc}
                \toprule
                \makecell[c]{Model\\@Dataset}& \makecell[c]{Non-IID\\Partition\\Strategy} & Method & Accuracy & \makecell[c]{Comm. \\Traffic} & \makecell[c]{Comm. Traffic \\Reduced to}\\
                \midrule
                \multirow{6}*{\makecell[c]{Logistic\\@FMNIST}} & \multirow{2}*{$\#C=2$} & STC & $81.75\% $ & \multirow{2}*{$0.21$MB} & \multirow{2}*{$0.14\% $} \\
                \cline{3-4}
                & & \textbf{$\gamma$-FedHT-Q} & $\mathbf{82.19}\% $ & & \\
                \cline{2-6}
                & \multirow{2}*{$\#C=3$} & STC & $82.73\% $ & \multirow{2}*{$0.20$MB} & \multirow{2}*{$0.13\% $} \\
                \cline{3-4}
                & & \textbf{$\gamma$-FedHT-Q} & $\mathbf{82.92}\% $ & & \\
                \cline{2-6}
                & \multirow{2}*{$\#C=5$} & STC & $ 83.36\% $ & \multirow{2}*{$0.16$MB} & \multirow{2}*{$0.10\%$} \\
                \cline{3-4}
                & & \textbf{$\gamma$-FedHT-Q} & $ \mathbf{83.46}\% $ & & \\
                \midrule
                
                
                \multirow{6}*{\makecell[c]{CNN\\@CIFAR-$10$}} & \multirow{2}*{$\#C=2$} & STC & $58.83\% $ & \multirow{2}*{$1.42$MB} & \multirow{2}*{$0.020\% $} \\
                \cline{3-4}
                & & \textbf{$\gamma$-FedHT-Q} & $\mathbf{63.90}\% $ & & \\
                \cline{2-6}
                & \multirow{2}*{$\#C=3$} & STC & $64.59\% $ & \multirow{2}*{$1.56$MB} & \multirow{2}*{$0.022\% $} \\
                \cline{3-4}
                & & \textbf{$\gamma$-FedHT-Q} & $\mathbf{71.26}\% $ & & \\
                \cline{2-6}
                & \multirow{2}*{$\#C=5$} & STC & $70.19\% $ & \multirow{2}*{$1.22$MB} & \multirow{2}*{$0.017\% $} \\
                \cline{3-4}
                & & \textbf{$\gamma$-FedHT-Q} & $\mathbf{74.10}\% $ & & \\
                \bottomrule
            \end{tabular}
        }
    }
    \vspace{-1em}
\end{table}

\subsection{Comparison of Communication Traffic}\label{V-C}

Experimental results show that $\gamma$-FedHT performs better than HT and Top-$k$ under equal communication traffic. In this part, we focus on two metrics, accuracy and communication traffic. We compare $\gamma$-FedHT with HT and Top-$k$ (under the same total communication traffic) in Table~\ref{table:CommW/oQ}. We compare $\gamma$-FedHT-Q with STC in Table~\ref{table:CommWQ}. A detailed discussion follows.

\noindent \textbf{$\gamma$-FedHT with vanilla HT:} $\gamma$-FedHT achieves higher accuracy with less communication traffic than HT under nearly all cases. This shows that the adaptive mechanism of our design effectively optimizes the training process of HT and avoids the waste of communication traffic. 

\noindent \textbf{$\gamma$-FedHT with Top-$k$:} In Logistic@FMNIST, $\gamma$-FedHT exhibits a higher accuracy by $0.26\%$ compared to Top-$k$ under $\#C=2$. In CNN@CIFAR-10, this accuracy difference expands to $7.42\%$ under $\#C=3$. This suggests that \textit{$\gamma$-FedHT can achieve better communication-accuracy trade-off than the SOTA sparsifier in FL, especially under non-convex and communication-constrained cases}.

\noindent \textbf{$\gamma$-FedHT-Q with STC:} Similar to Fig.~\ref{fig:experiment2}, the performance of $\gamma$-FedHT is not affected whether it carries a quantizer. Even with an extremely aggressive compression strategy, $\gamma$-FedHT does not introduce serious accuracy degradation, which validates the conclusion that $\gamma$-FedHT converges at the same rate as FedAVG.

\section{Related Works} \label{VI}


\noindent\textbf{FedAVG with gradient sparsification:} Research in this area has achieved impressive compression ratios as low as $1\%$ or less. However, many studies lack theoretical analysis, and the computational complexity cannot achieve $\mathcal{O}(d)$. One such study proposes STC \cite{stc}, which manages to achieve a compression ratio of  nearly $0.1\%$ without significant accuracy degradation. This is accomplished by implementing the downstream compression and encoding on Top-$k$ combined with ternary quantization. Other studies with similar approaches include FedZIP \cite{FedZIP} and B-MUSTC \cite{cui2021slashing}. The work \cite{infocom24FedGC} jointly considers adaptive node selection and sparsification compression, but does not derive the number of iterations required to converge to a specified error. The work \cite{Fed-EF} analyzes the convergence rate of FedAVG when using the sparsification compression, which is $\mathcal{O}(\frac{1}{\sqrt{T}})$ (for both the convex and non-convex scenarios), slower than vanilla FedAVG. $\gamma$-FedHT, however, can achieve the same asymptotic convergence rate as FedAVG and keep the low-cost feature.

\noindent\textbf{Low-cost compression in FedAVG:} 
Most works only use the quantization compression to keep the time complexity of $\mathcal{O}(d)$ or even less. They provide the convergence analysis, but typically only achieve a compression ratio of nearly $10\%$. One such work proposes \cite{cui2021slashing} MUSTC, an unbiased version of B-MUSTC which converges at the rate of $\mathcal{O}(\frac{1}{T})$ under convex scenarios. Similar works combine the quantization compression with mechanisms such as periodic aggregation \cite{FedPaQ}, downstream compression \cite{QuantityGlobal} and local gradient tracking \cite{FedCOM}. The work \cite{cui2022infocom} proposes an adaptation framework for robust dynamic networks by strategically adjusting the compression ratio, but not considers EF. The work \cite{cepe-FL} proposes Cepe-FL, propose a two-way adaptive compressive sensing scheme in FL and reduce the computational complexity from $\mathcal{O}(n)$ to $\mathcal{O}(1)$, but not guarantee the model convergence. The work \cite{FedComp} reduces the computational cost of Top-$k$ by compressing the indexes of compressed parameters and proposes FedComp, but still has the GPU-unfriendly operation. 



\noindent\textbf{Theoretical analysis of sparsification compression in non-IID scenarios:} 
Most works ignore the node selection, infrequent communication, and the decaying learning rate, thus not applicable for FL. The work \cite{stich2020communication} compares distributed quantized SGD with unbiased quantizers and distributed SGD with Error-Feedback and biased compressors in non-IID scenarios. The work \cite{dagc} proposes DAGC, which assigns compression ratios according to the training weight. The work \cite{EF21} proposes EF21, which refines the traditional error-feedback mechanism. The work \cite{CFedAVG} introduces a compression-based FL algorithm equipped with EF and achieves the same convergence rate as vanilla FedAVG in the non-convex cases, but lacks the analysis in the convex cases and does not consider the absolute compressor. 
Our theoretical analysis considers both the infrequent communication and the partial node participation, making it suitable for FL.


\section{Conclusion} \label{conclusion}

In this paper, we propose an ideal sparsifier for FL with a time complexity of $\mathcal{O}(d)$, named \( \gamma\)-FedHT. We first reveal that the hard-threshold compressor induces accuracy degradation in FL and the decaying-$\gamma$ in non-IID scenarios leads to the failure of this compressor in FL. Then, we propose $\gamma$-FedHT, a stepsize-aware low-cost hard-threshold compressor in FL, with the time complexity of $\mathcal{O}(d)$ and the same convergence rate as FedAVG. Experimental results show that $\gamma$-FedHT can improve accuracy by up to $7.42\%$ over Top-$k$ under the equal communication amount in non-IID scenarios. $\gamma$-FedHT is expected to replace Top-$k$ as the SOTA sparsifier in FL due to its excellent performance.



\clearpage
\renewcommand*{\bibfont}{\footnotesize}
\printbibliography

\clearpage
\end{document}